\documentclass[10pt,twoside,a4paper]{IEEEtran}

\usepackage{array}
\usepackage{graphicx}
\graphicspath{ {./images/} }

\usepackage[english]{babel}
\usepackage{hyperref}
\usepackage{verbatim}
\usepackage{mathtools} 
\usepackage{amsmath}
\usepackage{amssymb}
\usepackage[T1]{fontenc}
\usepackage[utf8]{inputenc}

\providecommand{\keywords}[1]
{
  \small	
  \textbf{\textit{Keywords:}} #1
}

\usepackage{cite}

\IEEEoverridecommandlockouts{}

\begin{document}
%
\renewcommand{\thepage}{A-\arabic{page}}

%
\title{Unsupervised Change Point Detection for heterogeneous sensor signals

}

\author{
\IEEEauthorblockN{Mario Krause\\}
\IEEEauthorblockA{Friedrich-Alexander Universität Erlangen-Nürnberg}
}

\maketitle

\begin{abstract}




Change point detection is a crucial aspect of analyzing time series data, as the presence of a change point indicates an abrupt and significant change in the process generating the data. While many algorithms for the problem of change point detection have been developed over time, it can be challenging to select the appropriate algorithm for a specific problem. The choice of the algorithm heavily depends on the nature of the problem and the underlying data source. In this paper, we will exclusively examine unsupervised techniques due to their flexibility in the application to various data sources without the requirement for abundant annotated training data and the re-calibration of the model. The examined methods will be introduced and evaluated based on several criteria to compare the algorithms. 
\end{abstract}

\keywords{Change point detection, Segmentation, Time series data}

\section{Introduction}
In the modern world a large amount of data is generated by all the interconnected devices and networks and collected via sensors \cite{Comparative_study_on_CPD}. These sequences of measurements from the sensors are indexed with a time stamp and called time series data. This massive amount of data can not be monitored without automation, so time series analysis has become increasingly important. In the context of change point detection, time series analysis is used to detect changes in the behaviour of the system, which can be due to internal or external properties. The task of identifying a specific point in time where the statistical properties of the underlying model of a signal or time series change is called change point detection. \cite{aminikhanghahisurvey} \\
\\
Change point detection is an essential topic in time series analysis with a wide range of applications that require the detection of abrupt changes in the data, for example to indicate a transition of the system from one state to another or to indirectly detect the state of a system by estimating that the state changed \cite{Comparative_study_on_CPD}. Here are some examples of applications that use change point detection: \\
\\
\textbf{Climate change detection} \\
Climate analysis and climate monitoring have become very important over the last few decades due to the possible occurrence of climate change caused by the increase in greenhouse gases in the atmosphere. In this context change point detection is used to discover climatic discontinuities and changes in the temperature. \cite{climate_change, Climate_AReviewandComparisonofChangepointDetectionTechniquesforClimateData , climate} \\
\\
\textbf{Medical condition monitoring} \\
Change point detection methods can be used to monitor patients health data and quickly detect abrupt changes of the medical conditions. These algorithms can be used to analyze changes in physical activity during the strength training program \cite{Medical_Condition_Strengh_Training} or to monitor the heart rate of patients undergoing anesthesia and surgery to indicate disturbances to the cardiovascular system \cite{Medical_Heart_Rate_Monitoring}. \\
\\
\textbf{Stock Market Analysis} \\
In the stock market the fluctuation of any stock price is normal according to economic theory, but there are some shifts that are abnormal and worth the investors special attention \cite{chen2012parametric}. In momentum strategies it is necessary to identify momentum turning points, when a trend reverses from an uptrend to a downtrend such as in the 2020 market crash due to the covid outbreak \cite{Stock_Market}. \\
\\
This article presents an overview and comparison of algorithms commonly used for detecting change points in time series data. The focus is on unsupervised change point detection, which involves segmenting the data without relying on large amounts of annotated training data or the need to re-calibrate the model for each data source. The goal of this article is to help choosing the right detection method for a particular application, with an emphasis on practical aspects like the implementation and the calibration of the parameters. Our selection of methods aims for a good general performance for different data sources without fine tuning the algorithm. In practice fine tuning a method for each sensor will most probably yield better performance, but adds significant overhead if the system has multiple heterogeneous signals with the number of signals possibly growing in the future. We are especially focusing on methods that can be applied to different heterogeneous sensor signals, like temperature or pressure measurements, without the requirement to tune the parameters separately for each data source. The methods should be able to detect change points in different 1-dimensional sensor signals from a complex heterogeneous system to discover dependencies between the signals. 
\section{Background}
\label{Background}
\begin{figure}[h!]
\centering
\includegraphics[scale=0.25]{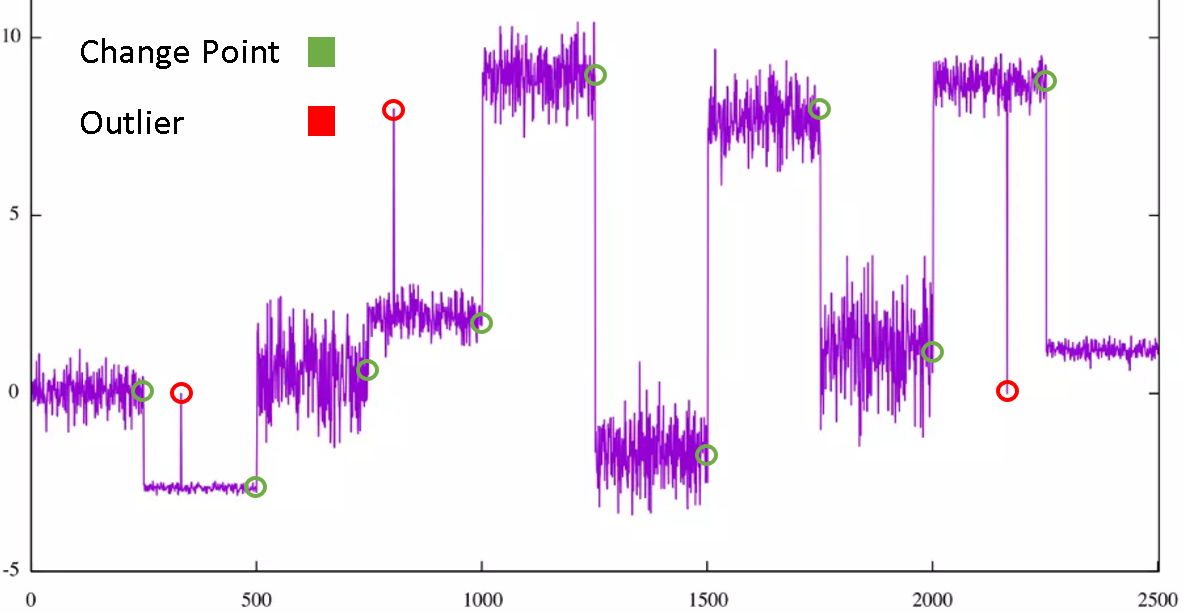}
\caption{Time series with change points and outliers}
\label{fig:CPD_Outlier}
\end{figure}

A change point occurs when the statistical properties of the time series data change abruptly. This change of measurements can be caused, for example, by a transition in the state of the underlying data generating system. These changes may be indicated by a simple shift in mean or variance, or more complex shifts such as in the frequency domain. An abrupt change refers to any alteration in the system's parameters that happens either instantly or very quickly relative to the measurement sampling period. It's important to note that abrupt changes don't necessarily indicate large shifts. In fact, detecting small changes is often the main challenge in most applications. \cite{Detection_of_Abrupt_Changes_CUSUM} \\
\\
In time series analysis mainly two concepts are used to detect deviations from the normal behavior in a data set: outlier detection and change point detection. The outlier detection is a method used to identify data points that are significantly different from the rest of the data, but are mainly very short shifts in the system or measurement errors. In contrast to that, change point detection aims to discover points in time where the state of the data generator changes for a subsequent and longer period. Therefore they are related, but not entirely similar. Figure \ref{fig:CPD_Outlier} illustrates the difference between these concepts by showing an example time series that contains several change points, where the mean and variance of the time series switches abruptly, and outliers, where an individual observation is far from the majority of the data. The main underlying assumption when investigating change points is that the properties or parameters describing the data are either constant or slowly changing over time in each state of the system \cite{Detection_of_Abrupt_Changes_CUSUM}. 

\subsection{Offline and online algorithms}
Change point detection algorithms are traditionally divided into two main branches: online and offline methods. Although both classes of algorithms aim to detect a change point in time series data, their methods of operation differ substantially. The online change point detection algorithms operate in real-time and run concurrently with the process they are monitoring, processing each data point as it becomes available. The constraint for these algorithms is that the processing should be completed before the arrival of the next data point \cite{A_novel_changepoint_detection_algorithm}. These methods are typically used in embedded applications such as monitoring industrial processes. Online change point detection methods must be able to process data quickly and make decisions in near real time. However, it is worth noting that online methods can also be applied in an offline setting, by analyzing stored data in a sequence. \\
\\
In contrast to that are the offline algorithms, which retrospectively detect changes and considers the entire data set. These methods can also be used in an online setting by collecting a small batch of data and detecting the change points in this batch. In general the concept of online and offline methods is a spectrum describing the number of data points the algorithm have to look at ahead of a potential change point, with complete offline methods considering the entire data set and complete online methods operating in perfect real-time. The offline methods allow the detection of multiple change points at once and focus more on the accuracy of the detection, while the online methods try to detect the most recent change point as fast as possible. Offline change point algorithms are typically more computationally intensive than online algorithms, as they have access to the entire data set and can use more complex methods to detect change points \cite{aminikhanghahisurvey}. The choice between online and offline methods should be based on the specific requirements of the application, with the most important consideration being the real-time constraint. Since the motivation for this paper is an application in an offline setting with the entire data set available, we will cover online and offline methods. 

\subsection{Supervised and unsupervised algorithms}
A variety of supervised and unsupervised methods have been developed and improved over time to tackle the challenge of change point detection. The supervised methods utilize machine learning algorithms that are trained on labeled data, where the change point locations are already known. The objective of these supervised algorithms is to build a model that can learn from the limited labeled data and accurately predict the change points for the entire data set. A variety of supervised methods can be used for this learning problem, such as decision tree, naive Bayes, support vector machines, logistic regression and nearest neighbor methods \cite{aminikhanghahisurvey}. One of the main drawbacks of supervised change point detection methods is the need for labeled data, as well as the need to retrain the model for each individual data source. This can make the process of applying these supervised methods to different data sources very costly and time consuming. \\
\\
Unsupervised learning algorithms are better suited for this as they are used to identify patterns in unlabeled data. In the context of change point detection, these algorithms can divide time series data into segments, with one segment before the change point and one after. These algorithms utilise statistical aspects of the data to locate change points in the time series. Unsupervised change point detection is attractive because it can handle a variety of different data sets without requiring prior training for each data set \cite{aminikhanghahisurvey}. Both types of methods have been shown to be effective, but for the purposes of this paper, we will focus solely on unsupervised methods. This decision is based on the underlying application of this paper that requires methods that are easily applicable to different data sources without retraining the algorithm. For an overview on supervised methods, we refer readers to \cite{Comparative_study_on_CPD, aminikhanghahisurvey}.

\subsection{Stability}
An important criteria for our application is the stability of the algorithms in terms of their ability to detect change points in different data sources without requiring prior fine tuning of the parameters for each data source. We are comparing the number of parameters and evaluate the effort of calibrating the algorithm. The parameters should be robust to a wide variation of values to minimize the effort of the parameter tuning. An important aspect for the stability of the algorithms is also the assumptions the algorithm makes about the underlying data source. \\
\\
In order to increase the stability of the algorithms, we are especially looking for algorithms that make little to no assumptions, therefor we are comparing the change point detection algorithms by determining if they are parametric or non parametric. It is important to make this distinction, because parametric models make assumptions about the underlying distribution of the data, such as for example the in the real world often appearing and therefor commonly used normal distribution. The distribution is described by a set of parameters, which have to be estimated from the data. The main issue of parametric models is that the assumptions about the underlying distribution may not hold for all data sources, which can lead to inaccurate estimates of the change points. The parametric nature of the algorithm can also be a benefit and lead to good results if the underlying distribution of the data is known. In contrast, the non-parametric models do not make assumptions about the underlying distribution of the data, which makes them more robust to different data sources. They skip the step of parameter estimation by using a non parametric approach by for example directly estimating the ratio between to densities without calculating the densities itself. \cite{aminikhanghahisurvey}

\subsection{Algorithm constraints}
Change point detection techniques can be differentiated based on the requirements imposed on the input data and the algorithm itself. These limitations play a crucial role in determining the most suitable method for detecting change points in a specific time series. Especially in our application it is important to have as little constraints as possible to ensure the algorithm can be applied to multiple data sources. Some algorithms for example are constraint to stationary or independent and identically distributed (i.i.d.) data sets. Another important constraint is the number of change points an algorithm can detect and if the number of change points has to be known prior to the run of the algorithm. Based on that we are classifying the algorithms into single and multiple change point detection algorithms. Algorithms that require the number of change points to be specified before the run will not be considered, since this is a major drawback when analysing different time series with an unknown amount of change points. The algorithms also differ in their outputs, with some algorithms providing the probability of a change point at a specific time point and others only returning the change points. The different outputs determine how easy the results can be interpreted and can for example give insights in the confidence of the detected change point. 

\subsection{Scalability}
The amount of time series data generated in the world is rapidly increasing, both in terms of the number of data points and dimensions. To effectively handle this massive amount of data, change point detection methods must be designed to be computationally efficient. Therefore, it is crucial to compare the computational cost of different change point detection algorithms and determine which methods can reach an optimal or approximate solution as quickly as possible. The problem of detecting change points in high dimensional data sets is not discussed in this paper, since our application is dealing with univariant time series. The computational costs of the algorithms are compared based on the information provided by the authors or estimated based on the algorithmic description. 

\section{Review}
This section provides an overview of commonly used change point detection algorithms. We will outline the basic principles of the algorithm and evaluate them based on the in section \ref{Background} defined criterion. These algorithms include both online and offline methods and where chosen based on their stability, algorithm constraints and scalability. All of the algorithms presented are unsupervised learning methods. 

\subsection{Likelihood ratio methods}
A typical statistical formulation of change point detection is to compare the probability distributions $p_{\theta_0}$ and $p_{\theta_1}$ of the data before and after a potential change point $x_t$. If these distributions are significantly different, the time point is considered a change point. One common method involves monitoring the logarithm of the likelihood ratio between consecutive intervals in the time series data. \cite{CPD_direct_density_ratio_estimation}
\begin{align*}
    s(x_i) \text{ = } \ln \frac{p_{\theta_1}(x_i)}{p_{\theta_0}(x_i)}
\end{align*}
We will discuss two different approaches that utilize the likelihood ratio in change point detection. The first approach is called the cumulative sum algorithm and relies on predesigned parametric models, where the probability density of the two consecutive intervals is calculated separately from the underlying data and the density ratio is calculated. The second approach is a direct density ratio estimation method, which is a non parametric algorithm where the ratio of the probability densities is directly estimated, without the requirement to perform density estimation for the individual segments. \\

\subsubsection{CUSUM}
\label{cusum}
On of the most familiar change point detection algorithms is the cumulative sum algorithm (CUSUM). Page \cite{Page} was the first to suggest the use of a cumulative sum to find changes in a parameter of interest. The CUSUM algorithm monitors the cumulative sum of the differences between successive measurements and a reference value or target. If this cumulative sum exceeds a predetermined threshold, it indicates that there has been a change point in the process. The log likelihood ratio is a measure of the difference between two distributions, and can be used to determine whether a change point has occurred. \\
\\
The typical behavior of the log likelihood ratio corresponds to a negative drift before change, and a positive drift after change. Therefore, the relevant information for change point detection lies in the difference between the value of the log-likelihood ratio and its current minimum value \cite{Detection_of_Abrupt_Changes_CUSUM}. Assume we have a sequence ${x_1, x_2,...}$ of time series variables with a unknown change point $x_tt$ and the distribution $p_{\theta_0}(x_t)$ before and $p_{\theta_1}(x_t)$ after the change point $x_t$. Then the corresponding decision rule for the CUSUM algorithm is:

\begin{equation}
    S_n \text{ = } \sum_{i=1}^{n} \ln \frac{p_{\theta_1}(x_i)}{p_{\theta_0}(x_i)} - \min_{k \leq n} \sum_{i=1}^{k} \ln \frac{p_{\theta_1}(x_i)}{p_{\theta_0}(x_i)} > L\\
\end{equation}
The CUSUM algorithms detects a change point when the CUSUM statistic $S_n$ is larger then the predefined threshold $L$. Another common approach, that is computationally more efficient, is to calculate the CUSUM statistic $S_n$ recursively in the following way:
\begin{equation}
    S_n \text{ = } \max \left( 0, S_{n-1} + \ln \frac{p_{\theta_1}(x_i)}{p_{\theta_0}(x_i)} \right)
\end{equation}
In practice, the CUSUM algorithm is implemented by first setting an initial value for the CUSUM statistic $S_0$, usually set to zero. Then the algorithm iterates through the observations, updating the CUSUM statistic at each time step. The algorithm continues until a change point is detected or the end of the data is reached. \\
\\
\textbf{Stability}: The CUSUM algorithm is parametric and makes assumptions about the underlying distribution of the data. In each iteration of the algorithm the set of parameters describing the distribution before a potential change point $p_{\theta_0}$ and after the change point $p_{\theta_1}$ have to be estimated from the data. A common application is when the process is normally distributed with the pre-change and post-change distributions having the same known variance $\sigma$. Then the interest centers on detecting shifts away from the pre-shift mean $\mu_0$ to $\mu_1$. The algorithm can handle different types of distributions, but if the assumption of the underlying distribution is not correct the result of the algorithm may vary.  \cite{Cusum_techniques} \\
\\
The algorithm has only two parameters, with the initial value of the CUSUM statistic usually set to zero and the threshold value $L$ which is more difficult to calibrate. The threshold value depends on the specific data set and the goals of the analysis, for example it can be set such that the false positive rate or the false negative rate is minimized or determined using cross-validation techniques. Different methods to calculate values of $L$ have been developed with a constant or adaptive threshold. A comparison of these methods is given in \cite{CUSUM_Thresholds}. \\
\\
\textbf{Algorithm constraints}: The standard CUSUM algorithm is an online algorithm for single change point detection, but many variations of this concept were developed over time for online and offline settings, as well as adaptions to multiple change point detection by running the algorithm with a sliding window. An overview of four of these adaptions is provided in \cite{Detection_of_Abrupt_Changes_CUSUM}. The algorithm has no constraints or limitations to the data it can be applied to wide variety of problems. The algorithm does not provide any information about the confidence of the output, it stops when a single change point is detected, indicated by the CUSUM statistic exceeding the threshold, and directly outputs the change point. \\
\\
\textbf{Scalability}: The main advantage of CUSUM is the simplicity of the underlying concept, which makes it easy to implement. The computational complexity of the CUSUM algorithm is generally considered to be low, as it only involves simple mathematical operations such as addition. The complexity estimated based on the algorithm is $O(n)$ for a sequence of n points. \\

\subsubsection{KLIEP}
The Kullback–Leibler importance estimation procedure (KLIEP) is the non parametric counter part to the CUSUM algorithm and also uses the likelihood ratio to detect change points. KLIEP avoids the problems that come with the parametric model of the CUSUM algorithm by using a more flexible non parametric model. This grants the advantage that it does not rely on strong model assumptions. A naive method would be to use a non parametric approach to estimate the densities separately and use them to calculate the ratio, but this approach is ineffective due to the challenges of non parametric density estimation. Instead of individually estimating each density, we can directly estimate the density ratio. Different online methods have been developed that use the idea of direct density ratio estimation \cite{CPD_direct_density_ratio_estimation, CPD_by_relative_density_ratio_estimation, Hushchyn2020Direct_Density_Ratio_Estimation}. In these methods the density ratio between two consequent intervals $X$ and $X'$ is modeled by a non parametric Gaussian kernel:
\begin{equation}
\hat{w}(X) \text{ = } \frac{p(X)}{p(X')} \text{ = } \sum_{l=1}^{n} \alpha_l K_\sigma(X,X_l)
\label{eqn:somelabel}
\end{equation}
where $p(X)$ is the probability distribution of the interval $X$ and the parameters $\alpha$ are learned from data samples. $K_\sigma (Y ,Y')$ is the Gaussian kernel function with mean $Y'$ and standard deviation $\sigma$. The parameters $\alpha$ are determined in a training phase by minimizing a dissimilarity measure. From the density ratio estimator $\hat{w}(X)$ an approximation of the dissimilarity measure between two samples is calculated and the higher the dissimilarity measure is, the more likely the point is a change point \cite{Hushchyn2020Direct_Density_Ratio_Estimation}. One of these methods is the Kullback-Leibler importance estimation procedure (KLIEP) that estimates the density ratio by using Kullback-Leibler (KL) divergence:
\begin{align*}
    KL [p(x)||p'(x)] \text{ = }& - \int p'(x) \log \frac{p(x)}{p'(x)} dx 
\end{align*}
The parameters $\alpha$ in the formula \ref{eqn:somelabel} are determined so that the empirical Kullback-Leibler divergence is minimized. The solution of this problem can be obtained by solving a convex optimization problem. With the estimated parameters, the logarithm of the likelihood ratio is evaluated as the change detection score and a change point is detected if the score is beyond a given threshold $L$
\begin{equation}
   S \text{ = } \sum_{i=1}^{n} \ln \hat{w}(X_i) > L
\end{equation}
\textbf{Stability}: The KLIEP algorithm has several tuning parameters, such as the kernel width $\sigma$, the lengths of the two intervals $X$ and $X'$ and the threshold for the change point score $L$. The main advantage of the method described in \cite{CPD_direct_density_ratio_estimation} is that it is equipped with a natural cross validation procedure for tuning the kernel width parameter $\sigma$. The interval length has an impact on the accuracy of the estimation of the density ratio, with larger intervals leading to better estimates but also to possible issues in data sources where change points occur very frequently. The detection of the change points is sensitive to the threshold parameter $L$, with possible methods to calibrate $L$ already discussed in chapter 1) CUSUM. \\
\\
The main advantage in terms of the stability of KLIEP compared to the CUSUM algorithm is the robustness to different types of data distributions. The KLIEP algorithm is non parametric and doesn't make assumptions about the underlying distribution of the data. This is especially useful if the underlying distribution of the data is unknown or the algorithm is applied to different data sources that do not necessarily have the same distribution.\\
\\
\textbf{Algorithm constraints}: The KLIEP and CUSUM algorithm share a lot of properties, with their sequential behaviour both are online methods use for single change point detection and can be adapted for multiple change point problems with small adjustments. The algorithm has no constraints or limitations to the data and directly returns the change point. \\
\\
\textbf{Scalability}: Compared with other approaches like the CUSUM algorithm, methods based on density
ratio estimation tend to be computationally more expensive because of the cross-validation procedure for model selection \cite{CPD_by_relative_density_ratio_estimation}. After the model selection the calculation of the change point score for a given density ratio is computationally very efficient and involves only simple mathematical operations that are performed in linear time. 

\subsection{Bayesian online change point detection}
\begin{figure}[h!]
\centering
\includegraphics[scale=0.25]{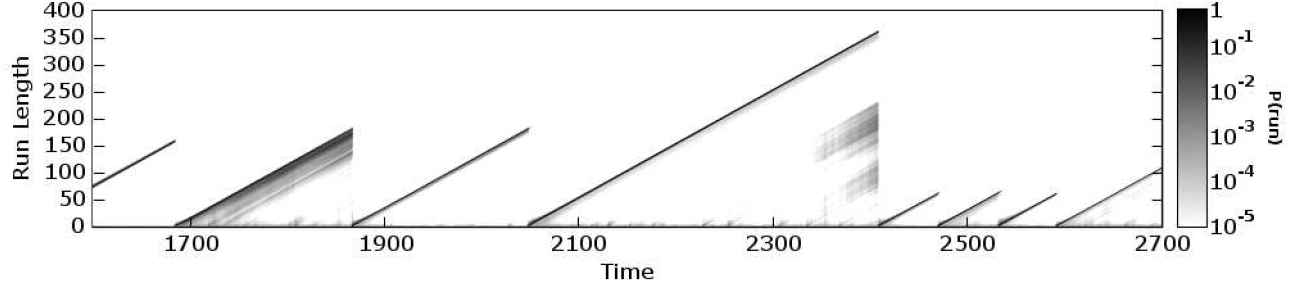}
\caption{Example of bayesian online change point detection from \cite{Bayesian_Online_Changepoint_Detection}}
\label{fig:BOCPD}
\end{figure}

A common bayesian method for detecting changes, described by Adam and MacKay \cite{Bayesian_Online_Changepoint_Detection}, involves estimating the posterior distribution of the run length $r_t$, which describes the time that passed since the last change point. The algorithm is an online method, so in each iteration a new data point is considered and the run length can either increase by 1 or drop down to 0. The set of observation $x_t^{(r)}$ is associated with the run length $r_t$ and includes all the observations of the current run, with new observations added to the set until a change point is found, which sets $r_t$ to 0 and $x_t^{(r)}$ empty. The algorithm determines the probability $P(r_t=0|x_{1:t})$ of a change point occurring at time $t$ by calculating the probability of the run length of 0 at time $t$ given the set of observations $x_{1:t}={x_1,...,x_t}$. The functionality of the algorithm is visualized in figure \ref{fig:BOCPD} with the run length and the change point probability represented as a grey scale, with darker pixels indicating a higher probability. The plot shows that the run length drops to zero when the probability of a change point gets close to one. 

This probability of a change point can be calculated based on Bayes theorem as:
\begin{align*}
    P(r_t | x_{1:t})=\frac{\sum_{r_{t-1}} P(r_t| r_{t-1})P(x_t|r_{t-1}, x_t^{r})P(r_{t-1}, x_{1:t-1})}{P(x_{1:t})}
\end{align*}
Since $P(r_{t-1}|x_{t-1})$ is a recursive component, it is known from the previous step and we only need to calculate the conditional run length probability $P(r_t | r_{t-1})$ and evaluate the predictive distribution $P(x_t|r_{t-1}, x_t^{r})$. The conditional prior on the change point $P(r_t| r_{t-1})$ is only nonzero at the two outcomes $r_t$=0 or $r_t$=$r_{t-1}$, which gives the algorithms its computational efficiency. \cite{Bayesian_Online_Changepoint_Detection} 
\begin{equation*}
    P(r_t|r_{r-1})=\left\{%
\begin{array}{ll}
    1 - H(r_{t-1}+1), & \hbox{if }r_t = r_{t-1}+1 \\
    H(r_{t-1}+1), & \hbox{if }r_t=0 \\
    0, & \hbox{otherwise} \\
\end{array}
\right.
\end{equation*}
The function H(r) is called the hazard function and describes how likely a change point is occurring at a run length r. A common approach is to make this process memoryless by setting H(r) = $\frac{1}{\lambda}$ with a timescale parameter $\lambda$. Another option would be to make H(r) increasing over the run to penalize longer run lengths. The predictive distribution $P(x_t|r_{t-1}, x_t^{r})$ represents the probability that the most recent observation belongs to current run and is the most challenging to calculate. In \cite{Bayesian_Online_Changepoint_Detection} it was proposed to use a conjugate exponential model with parameters $\nu$ and $\chi$ to make the algorithm computationally efficient. \\
\\
\textbf{Stability}
The bayesian online change point detection algorithm is a parametric method. The method allows us to encode knowledge about the world into the algorithm by setting the two conjugate priors $\chi$ and $\nu$. The conjugate priors $\chi$ and $\nu$ are the only two parameters of the algorithm and are set based on prior knowledge. This prior knowledge could be the mean and variance of the data, which is estimated roughly based on the first few observations. A possible option for further tuning the algorithm it by setting the hazard function H(r), but the common memoryless approach is sufficient for most applications. \\
\\
\textbf{Algorithm constraints}: The algorithm is a online multiple change point detection method that assumes the sequence of observations can be segmented into non-overlapping partitions and the data within each partition p is i.i.d. from some probability distribution $P(x_t |\mu_p)$ \cite{Bayesian_Online_Changepoint_Detection}. The restriction to i.i.d. time series is a major drawback for this algorithm when applying it to multiple different data sources. The algorithm directly returns the probability for a change point occurring at the specific run lengths, which makes the results easier to interpret compared to methods like CUSUM or KLIEP that only return the change point itself. \\
\\
\textbf{Scalability}: The implementation of the algorithm proposed by Adam and MacKay \cite{Bayesian_Online_Changepoint_Detection} is quadratic in space and time complexity in the number of data points $n$ so far observed. Clearly this is problematic when analysing long time series. To overcome this issue a simple approximation was introduced in \cite{malladi2013online} that reduced the complexity to $O(n)$.

\subsection{Singular spectrum transformation}
\begin{figure}[h]
\centering
\includegraphics[scale=0.7]{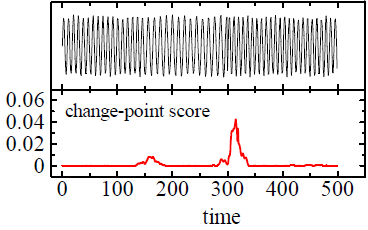}
\caption{Example of a time series and the corresponding change point score from \cite{SST2}}
\label{fig:SST}
\end{figure}

The singular spectrum transformation (SST) algorithm first proposed by Moskvina and Zhigljavsky \cite{SSA} uses singular spectrum analysis and adapts it for change point detection. The singular spectrum transform is a non parametric approach and can be used to analyze time series with complex structure \cite{SSA}. This technique involves transforming the original time series into a new series of change point scores. The resulting time series can be interpreted as the probability distribution that some change point occurs at time $t$. In figure \ref{fig:SST} such a change point score with the corresponding time series is shown with a clear peak in the change point score when the frequency of the time series changed. \\
\\
The underlying idea is to compute for each time point $x_t$ the difference between a representative pattern of a few time points before and a few points after $x_t$. The dynamics of the time series are represented using a Hankel matrix. We call the Hankel matrix, representing the change patterns within the past $w$ points, trajectory matrix $H(t)$ and the Hankel matrix for representing the future change patterns the test matrix $G(t)$. The representative pattern of $H(t)$ and $G(t)$ can be extracted by performing a singular value decomposition on both matrices. The $l < w$ left singular vectors $U_{i_1}(t), ... ,U_{i_l}(t)$ of $H(t)$ with the largest singular values represent the past change pattern as a hyperplane and build the matrix $U(t) \text{ = }U_{i_1}(t), ... ,U_{i_l}(t)$. The matrix $U(t)$ encodes the major direction of change in the past signal and the parameter $l$ is the number of representative patterns that are considered. The importance of each representative pattern $U_{i}(t)$ is given by the corresponding singular value, with the most dominant pattern corresponding to the largest singular value.  The direction of maximum change in the future of the signal is given by the left singular vector $\beta(t)$ of $G(t)$ with the largest singular value. \cite{SST2}\\
\\
To estimate the difference between the past and the future patterns, $\beta(t)$ will be projected onto $U(t)$ and normalized to calculate a change point score. If there is no change in the dynamics of the signal, it is expected that $\beta(t)$ will lie in or very near to the hyperplane represented by $U_l$. The change point score ranges from zero to one, with a high likelihood of a change point occurring when the score is close to one. The score can be calculated at any time point $t$ by finding representative patterns in both the trajectory and test matrices. This can be seen as a transformation from the original time series $T$ to a new time-series $T_c$. This demonstrates that SST can make variables of different types comparable by converting a heterogeneous system into a homogeneous one \cite{SST2}, i.e.
\begin{equation}
    T \rightarrow T_c(w,l,g,m,n)
\end{equation}
\textbf{Stability}: The main problem of the SST algorithm is the need to specify five parameters: the length of the column vectors of the Hankel matrix $w$, the number of columns of H(t) $n$, the number of singular vectors $l$, the shift of the starting point for the future signal $g$ and the number of columns of G(t) $m$. In \cite{SST2, robustSSA} it was shown that SST is usually robust to a wide range in different values of $w$ and $n$ and domain knowledge or visualization can help finding an appropriate value for both. The problem lies in choosing the other three parameters, where domain knowledge is not very useful. Despite the problem of specifying five different parameters it was shown in \cite{SST2} that for a wide range of $w$ (6 < $w$ < 40), the results of the algorithms with $w=-g=m=n$ and $l=3$ are quite robust and the essential features remain unchanged. \\
\\
\textbf{Algorithm constraints}: The SST is based on the singular value decomposition of two Hankel matrix, so it is a non parametric method, as it does not rely on any specific assumptions about the distribution of the data, making it suitable for a wide range of time series data \cite{SSA}. The advantage of the SST algorithms lies in its comparability of different time series. The algorithms returns a change point score for each time point, which allows to detect multiple change points in one run. The change point score can be interpreted as the probability that a change point occurred at a specific time point, which makes it easy to interpret the output of the algorithm and discover dependencies within multiple different heterogeneous time series. \cite{SST2} \\
\\
\textbf{Scalability}: The complexity of the singular spectrum transformation is linear in the length of the time series \cite{robustSSA}. In each of the $n$ time steps a singular value decomposition of the fixed size Hankel matrices $H(t)$ and $G(t)$ have to be computed, which makes SST less computational efficient as other change point detection algorithms with linear complexity. 

\subsection{Binary Segmentation}
\begin{figure}[h]
\centering
\includegraphics[scale=0.3]{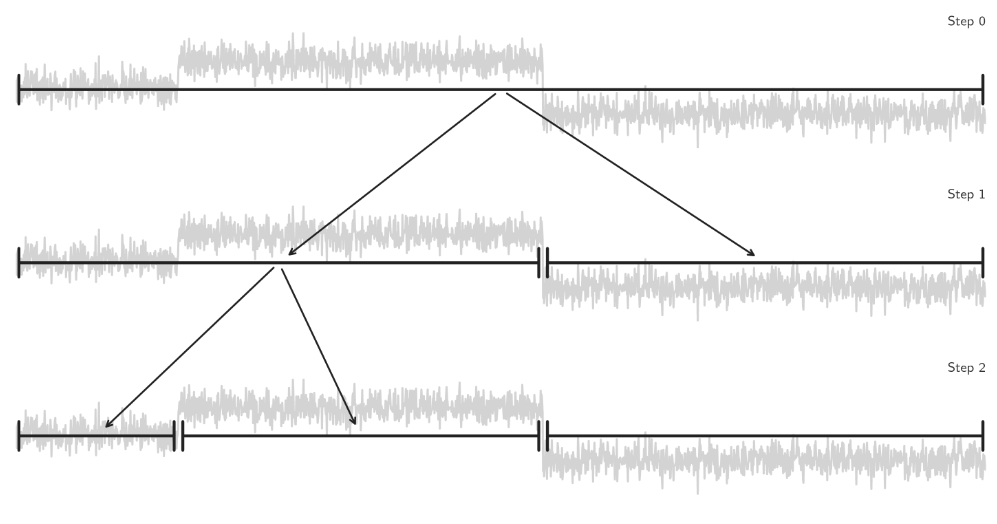}
\caption{Schematic example of binary segmentation from \cite{truong2020selective}}
\label{fig:BinSeg}
\end{figure}

Binary segmentation is a common technique for offline change point detection in time series data due to its simplicity both in the underlying concept and the implementation of the algorithm. A implementation of the algorithm available in the ruptures python library for offline change point detection and is called BinSeg \cite{truong2020selective}. The binary segmentation procedure is one of the standard methods for detecting multiple change points by using a test for single change point detection \cite{lavielle2007adaptiveBinSeg}.  \\
\\
The algorithm is a recursive technique for multiple change point detection in which initially the entire data set is searched for one change point. Once a change point is detected, the data set is split into two subsegments, defined by the detected change point. The algorithm is then performed on each of the two subsegments resulting in further splits. This procedure continuous in each new split until a stopping criteria is satisfied or no change point is detected. The indices of the segment boundaries are the change points. This process of recursively splitting the segments into two smaller subsegments is shown in the schematic view of the algorithm in figure \ref{fig:BinSeg}. \\

This technique can be used with a variety of time series and is able to find changes in the mean, variance, or distribution of the data. Binary segmentation is considered a greedy algorithm, because it is performed sequentially, with each stage only visited once and depending on the previous ones. \cite{wild_binSeg}\\
\\
The algorithm starts by initializing the starting and ending indices $s$ and $e$ of the time series sequence, for the first run start $s = 1$ and end $e = n$, where $n$ is the number of data points. Then a cost function $c(x_{a...b})$ is defined to measure the cost of a segment $x_{a...b}$, with $x_{a...b} = {x_i | a <= i <= b}$. The cost function should be chosen based on the nature of the data and the goal of the analysis. An overview and comparison of common cost functions is given in \cite{truong2020selective}. The algorithm calculates the costs for each possible split by summing the cost function for the two resulting segment and returns the change point $\hat{k}$ that minimizes the costs:
\begin{equation}
    \hat{k} = \arg \min_{s<k<e} c(x_{s...k}) + c(x_{k...e})
\end{equation}
The algorithm then recursively calls the binary segmentation algorithm with the new start and end indices: BinSeg($s$, $k$) and BinSeg($k$, $e$). This process is repeated until a stopping criterion is met, such as a minimum number of data points in a segment, the maximum number of change points or a maximum number of iterations. Once the stopping criterion is met, the change points are assigned as the indices of the segment boundaries.\\
\\
\textbf{Stability}: The algorithm is very simple and has only the cost function and a stopping criteria as inputs. It's important to note that the choice of the cost function $c()$ and the stopping criterion are crucial to the performance of the algorithm. The stopping criterion can be chosen according to the number of change points expected in the data, the granularity of the segments, or a threshold of the cost function for a segment. In general, the choice of cost function and stopping criterion depends on the nature of the data and the goal of the analysis. \\
\\
The algorithm can be used in a parametric and non parametric setting by selecting the respective cost function, depending on the prior knowledge before performing the change point detection. The choice of the cost function is determining the assumption the algorithm makes about the data. In \cite{truong2020selective} an overview of possible parametric and non parametric cost functions is given with their underlying assumptions about the data. \\
\\
\textbf{Algorithm constraints}: Binary segmentation is a simple and versatile offline algorithm for change point detection, that can be applied to any time series without any limitations. The algorithm can be used in situations where the number of change points is known prior to the run but also for an unknown number of change points. However, the algorithm has a limitation in accurately detecting change points that are too close together and result in small segments. This can lead to inaccuracies in the results for time series, that frequently change between different states resulting in close change points. The binary segmentation algorithm directly returns the detected change points, with the most significant change point appearing first. This feature allows the user to stop the algorithm at any point in time and still obtain a meaningful result, making it an efficient tool for change point detection.\\
\\
\textbf{Scalability}: The benefits from the simplicity of binary segmentation includes the low complexity of the algorithm of $O(C n log n)$, where $n$ is the number of samples and $C$ the complexity of calling the considered cost function on one subsegment \cite{truong2020selective}. 

\subsection{Bottom up Segmentation}
\begin{figure}[h]
\centering
\includegraphics[scale=0.3]{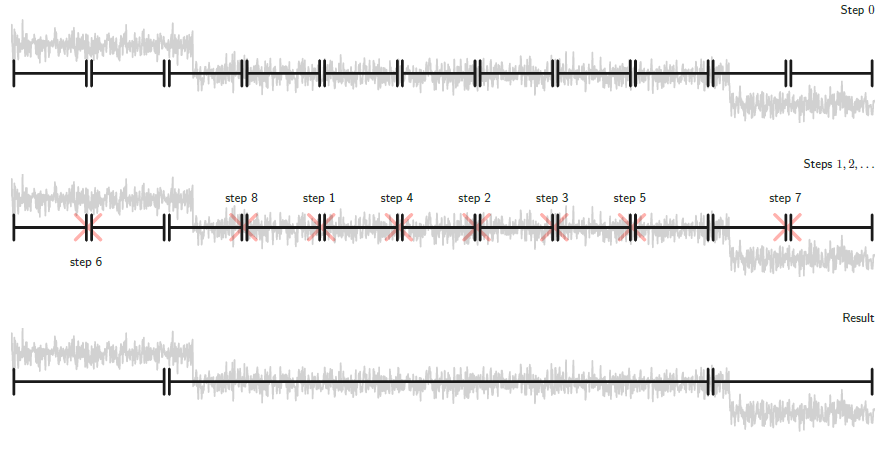}
\caption{Schematic example of bottom up segmentation from \cite{truong2020selective}}
\label{fig:BottomUp}
\end{figure}

Bottom-up segmentation is the natural counterpart of binary segmentation. In contrast to binary segmentation, bottom up segmentation starts by dividing the time series into individual observations, each treated as a separate segment, followed by sequentially merging adjacent segments based on a discrepancy criterion. A schematic overview of the bottom up segmentation is provided in figure \ref{fig:BottomUp}, showing the different steps of the algorithm starting with a segmented time series in a grid and merging these segments together until a stopping criterion is met and the approximation of the change points is returned. \\

Bottom up segmentation is like its counterpart binary segmentation available in the ruptures python library \cite{truong2020selective}. It is also an offline method with a very simple underlying concept and low computational complexity. \\
\\
The algorithm for bottom up segmentation starts by creating the finest possible representation of the time series of size $n$ by dividing it into $\frac{n}{2}$ segments. The indices of the segment boundaries are the potential change points. For each potential change point $t$ a discrepancy measure $d(x_{a...t}, x_{t...b})$ is calculated between the segments $x_{a...t}$ and $x_{t...b}$ separated by $t$, with these segments defined as $x_{a...b} = \{x_i | a \leq i \leq b\}$. For a given cost function $c()$, the discrepancy measure between two adjacent segments is given by:

\begin{equation}
    d(x_{a...t}, x_{t...b}) = c(x_{a...b}) - c(x_{a...t}) - c(x_{t...b})
\end{equation}

All potential change points are ranked by the discrepancy measure with the lowest discrepancy change point getting deleted, meaning that the corresponding segments are merged. 
When two adjacent segments $t$ and $t+1$ are merged, the algorithm performs some bookkeeping tasks and calculates the cost of merging the newly formed segment with its right neighbor segment. Additionally, the costs for merging the left neighbor with the newly formed larger segment have to be recalculated. These bookkeeping steps are necessary to ensure the accuracy of the bottom up segmentation process and its ability to detect change points in the time series. \cite{BinSEG_BottomUP}\\
\\
The process of merging and evaluating segments is repeated until no further improvements can be made. This results in a set of segments that represent the best approximation of the underlying process changes in the time series. The final step is to identify the change points, which correspond to the boundaries between the segments and represent the points where the underlying process changes.\\
\\
\textbf{Stability}: Due to the simple underlying concept of the bottom up algorithm the only two inputs are the cost function, used in the discrepancy measure, and a stopping criteria. The choice of the two inputs have a significant impact on the performance of the algorithm. The stopping criterion can be chosen according to the number of change points expected in the data, the size of the segments, or a threshold of the discrepancy measure for two segments. In general, the cost function and stopping criterion should be chosen based on the nature of the time series and the goal of the analysis. \\
\\
Similar to binary segmentation this algorithm can be parametric and non parametric, depending on the selected  cost function. The cost function should be chosen based on the prior knowledge of the data source before performing the change point detection. The choice of the cost function is determining the assumption the algorithm makes about the data. In \cite{truong2020selective} an overview of possible parametric and non parametric cost functions is given with their underlying assumptions. \\
\\
\textbf{Algorithm constraints}: The algorithm has no limitations in the time series it can be applied to and can be used for a known and unknown number of change points. Due to the very fine grid in the initialization the first iterations of merging procedures can be unstable because of the small segments they are performed on, for which the statistical significance is smaller. This problem could be overcome by starting with larger segments. However if a true change point does not belong to the original set of boarders of these segments, it would never be considers for a change point, resulting in an inaccurate detection of the change points. \cite{truong2020selective}\\
\\
\textbf{Scalability}: The algorithm for bottom up segmentation has a similar complexity as their counterpart binary segmentation with $O(C$ $n$ $log(n))$, where $n$ is the number of samples in the time series and $C$ the complexity of calling the discrepancy measure for two adjacent subsegments \cite{BinSEG_BottomUP}. \\

\section{Discussion and comparison}
The previous sections present an overview of change point detection algorithms that are commonly used in the literature and have a good general performance for different data sources without fine tuning each algorithm for the particular data source. The task of selecting the algorithm best suited for a particular application can be challenging and depends on which criterion is most important for the application. We compare change point detection methods based on their stability, algorithm constraints and their scalability to help with this task. 

\subsection{Stability}
When evaluating the stability of algorithms, one important factor to consider is the number of parameters and the robustness of these parameters to various changes. While all algorithms except singular spectrum transformation have a small number of parameters, the choice of these parameters can have a significant impact on the performance of the algorithms. For instance, in the case of the CUSUM or KLIEP algorithm, only one threshold parameter $L$ needs to be assigned. However, the threshold value in CUSUM or KLIEP and the other stopping criterion in binary segmentation or bottom up segmentation can be very sensitive, and choosing an incorrect threshold value can result in either an early or a late stopping of the algorithm, which leads to inaccurate results. In contrast, singular spectrum transformation requires five different parameters to be assigned, but it has been shown to be effective in detecting change points in a variety of data sources without adjusting the parameters \cite{SST2}. Similarly, bayesian online change point detection algorithm only requires the specification of the conjugate priors as input, which is essentially a problem of estimating the distribution of the data. The absence of a stopping criteria in both singular spectrum transformation and bayesian online change point detection makes these algorithms relatively more stable compared to other algorithms. \\
\\
To enhance the stability of a change point detection method, it is important to consider the assumptions the algorithm makes about the data. In general, non parametric change point detection methods tend to be more robust than parametric methods, as the latter heavily depend on the choice of parameters used to model the distribution of the data \cite{aminikhanghahisurvey}. A naive but common approach is to use a parametric model and always assume that the different data sources have a normal distribution. Since this assumption is true for a lot of real world applications this approach may produce good results for data sets that actually follow a normal distribution, but it can lead to poor performance when the data deviates from a normal distribution. In contrast, non-parametric methods, such as KLIEP, binary segmentation, or bottom-up segmentation, do not make any assumptions about the underlying distribution of the data and are therefore more robust. A comprehensive overview of the algorithms and their parameters, along with a distinction between parametric and non-parametric methods, can be found in Table \ref{tab:AlgStab}.

\begin{table}[h]
\centering
\caption{Comparison of change point detection algorithms stability}
\label{tab:AlgStab} 
\begin{tabular}{ |p{1cm}||p{2cm}|p{4.5cm}|}
 \hline
 Method& Parametric/ \newline non parametric & Parameters \\
 \hline \hline
 Cusum      & Parametric                & threshold value $L$\\
 \hline
 KLIEP      & Non parametric            & kernel width $\sigma$ \newline lengths of the two intervals \newline threshold value $L$   \\
 \hline
 BOCPD      & Parametric                & conjugate priors $\chi$ and $\nu$ \\
 \hline
 SST        & Nonparametric             & length of the representative pattern w \newline number of columns of H(t) \newline number of singular vectors l \newline shift of the starting point g \newline number of columns of G(t)\\
 \hline
 BinSeg     & Parametric/ \newline non parametric & cost function \newline stopping criteria\\
 \hline
 Bottom Up  & Parametric/ \newline non parametric & cost function \newline stopping criteria\\
 \hline
\end{tabular}
\end{table}

\subsection{Algorithm Constraints}
When choosing the appropriate algorithm for a particular application, it is important to take into account the constraints of the algorithm. This review focuses on three main constraints: the restrictions on the time series, the number of change points that can be detected, and the output of the algorithm. The Likelihood ratio methods have no limitations on the time series, but they are only capable of detecting a single change point, so multiple runs of the algorithm may be required to detect multiple change points. This problem is overcome by directly using methods such as binary or bottom-up segmentation, which are designed for multiple change point detection and also do not have any limitations on the time series. Bayesian online change point detection has the only restriction on the time series in that it requires the time series to be independently and identically distributed. Singular Spectrum Transformation, on the other hand, requires the time series to be stationary. It is important to consider these constraints when selecting the right algorithm for a given application to ensure that the algorithm is capable of meeting the specific needs and requirements of the data. \\
\\
The various algorithms also have different outputs, with some only providing the change points and others providing a change point probability or score. The bayesian online change point detection and singular spectrum transformation algorithms provide a change point probability or score, but this information requires a separate post processing step to assign the actual change points. However, this additional information provides insight into the confidence of the algorithm in the identified change points. On the other hand, the likelihood ratio methods and binary and bottom up segmentation algorithms directly return the change points, avoiding the need for a separate post processing step. A comprehensive overview of the algorithms and their constraints can be found in table \ref{tab:AlgConst}. 

\begin{table}[h]
\centering
\caption{Comparison of change point detection algorithms constraints}
\label{tab:AlgConst} 
\begin{tabular}{ |p{1cm}||p{2cm}|p{2cm}|p{2cm}|}
 \hline
 Method& Time series \newline limitations & No. change points detected & Output \\
 \hline \hline
 Cusum      & No limitation             & Single & Change point\\
 \hline
 KLIEP      & No limitation            & Single & Change point\\
 \hline
 BOCPD      & i.i.d. time series        & Multiple & Change point probability\\
 \hline
 SST        & Time series should be stationary & Multiple& Change point score\\
 \hline
 BinSeg     & No limitation & Multiple & Change point\\
 \hline
 Bottom Up  & No limitation & Multiple & Change point\\
 \hline
\end{tabular}
\end{table}

\subsection{Scalability}
Another important criteria to consider are the computational costs of change point detection algorithms. Table \ref{complexity} presents a comparison of the computational cost of the algorithms surveyed. For some algorithms the computational costs where provided by the authors. In cases where the authors have not provided this information, the comparison was conducted based on the descriptions of the algorithms. \\
\\
The computational cost of CUSUM is relatively low, as it requires only a few calculations for each time step. The time complexity of CUSUM is linear in the number of data points $n$, making it a suitable option for large data sets. The KLIEP algorithm, on the other hand is a slightly more complex algorithm with a higher time complexity as CUSUM due to the model selection required. \\
\\
The computational cost of bayesian online change point detection depends on the choice of priors and the complexity of the model used. In general, it is computationally more expensive and has a quadratic time complexity $O(n^2)$, but this can be reduced to a linear complexity with the help of a simple approximation. \\
\\
The singular spectrum transform has also a linear time complexity but is less efficient as CUSUM or KLIEP as it requires the calculation of a singular value decomposition in each time step, which has a high time complexity. \\
\\
The computational complexity of binary segmentation and bottom up segmentation depends on the cost function used in the algorithm. Both have a complexity of $O(C$ $n$ $log(n))$, where $n$ is the number of samples in the time series and $C$ the complexity of calling the cost function. \\
\\
This review covered both online and offline change point detection methods. In general the online methods are computationally less expensive, because they focus on detecting the most recent change point as fast as possible. This was also confirmed by the algorithms in this review with two offline methods binary segmentation and bottom up segmentation having the highest computational complexity. 

\begin{table}[h]
\centering
\caption{Comparison of the algorithms computational complexity}
\label{complexity} 
\begin{tabular}{ |p{1cm}||p{3cm}|  p{3cm}|}
 \hline
 Method& Online / Offline method & Computational complexity \\
 \hline \hline
 Cusum      & online & O(n)*\\
 \hline
 KLIEP      & online &  Cusum < KLIEP*\\
 \hline
 BOCPD      & online & O(n)\\
 \hline
 SST        & online & KLIEP < SST* \\
 \hline
 BinSeg     & offline & $O(n \log n)$\\
 \hline
 Bottom Up  & offline  & $O(n \log n)$\\
 \hline
\end{tabular}
{\raggedright   *Estimated based on the algorithm \par}
\end{table}

\section{Conclusion}
In this review, we have analyzed various techniques for change point detection and organized them under a unified framework. Our emphasis was on unsupervised methods that have a good general performance for diverse data sources. To assess the methods, we compared them based on three criteria: stability, algorithmic constraints, and scalability. The aim is to provide a framework that can be utilized to assess future algorithms in a systematic manner. The focus of this review was chosen based on a particular application that was the motivation for this paper. It is worth noting that in practice fine tuning a specific algorithm for a single time series will most probably yield better performance, but significantly increases the time and effort needed to detect change points in a large system of heterogeneous signals. The comparison of the methods was based solely on the description of the algorithm and leaves room for further research by for example comparing the performance of the methods on different data sources and evaluating the results.




%

\bibliographystyle{plain}
\bibliography{bibliography.bib}

\end{document}